# Using Qualitative Hypotheses to Identify Inaccurate Data

**Qi Zhao** QI-ZHAO@IS.AIST-NARA.AC.JP
**Toyoaki Nishida** NISHIDA@IS.AIST-NARA.AC.JP
*Graduate School of Information Science*
*Nara Institute of Science and Technology*
*8916-5, Takayama-cho, Ikoma-shi, Nara 630-01, Japan*

## Abstract

Identifying inaccurate data has long been regarded as a significant and difficult problem in AI. In this paper, we present a new method for identifying inaccurate data on the basis of qualitative correlations among related data. First, we introduce the definitions of related data and qualitative correlations among related data. Then we put forward a new concept called support coefficient function ($SCF$). $SCF$ can be used to extract, represent, and calculate qualitative correlations among related data within a dataset. We propose an approach to determining dynamic shift intervals of inaccurate data, and an approach to calculating possibility of identifying inaccurate data, respectively. Both of the approaches are based on $SCF$. Finally we present an algorithm for identifying inaccurate data by using qualitative correlations among related data as confirmatory or disconfirmatory evidence. We have developed a practical system for interpreting infrared spectra by applying the method, and have fully tested the system against several hundred real spectra. The experimental results show that the method is significantly better than the conventional methods used in many similar systems.

## 1. Introduction

In many problems of artificial intelligence, inferences are drawn on the basis of interpretation or analysis of measured data. However, when measured data are inaccurate, interpreting or analyzing them is very difficult. In diagnosis or signal analysis, for example, the general reasoning method is to compare measured data with reference values (Reiter, 1987; Shortliffe & Buchanan, 1975). When measured data are not accurate due to noise or other unforeseen reasons, the comparison between measured data and reference values can not lead to any useful conclusion. A rule like "*if there is a strong peak in 3000 $cm^{-1}$- 3100 $cm^{-1}$ on the infrared spectrum of an unknown compound, then the unknown compound may contain at least one benzene-ring*" may work in ideal cases. However, the rule can not work in general cases. For example, when the spectral data are inaccurate, e.g., the measured peak in 3000 $cm^{-1}$- 3100 $cm^{-1}$ is not a strong peak but a medium one, or a measured strong peak is not exactly located in 3000 $cm^{-1}$- 3100 $cm^{-1}$ but is slightly shifted, the rule may not be applied.

In practical problems, especially in data rich problems such as diagnosis and interpretation, measured data are often inaccurate. One reason is that the measuring methods are error-prone. For example, a patient's temperature or blood-pressure may be inaccurately measured or entered, and a witness may inaccurately describe the features of a criminal. The other reason is that the real data are not noise-free. For example, among the received signals, there may be some noise mixed up, and what is worse, infrared spectral data (peaks) themselves may be noisy, i.e., some peaks may be affected by noise or other factors.

Identifying inaccurate data has long been regarded as a significant and difficult problem in AI. Many methods have been proposed to deal with the problem. Fuzzy logic provides a mathematical framework for representation and calculation of inaccurate data (Zadeh, 1978). By fuzzy logic, reference value $x_0$ is associated with a fuzzy interval $\triangle x$. If a measured data item falls into $[x_0 - \triangle x, x_0 + \triangle x]$, then it can be identified as the reference value with a corresponding membership degree. Probability theory and possibility theory are also widely used for handling inaccuracy and uncertainty (Dempster, 1968; Duda, Hart, & Nilsson, 1976; Pearl, 1987; Shafer, 1976; Shortliffe & Buchanan, 1975). The above methods are commonly used in AI systems. The way of applying them, however, depends on the nature of domain problems, and there is not yet a standard and generally accepted method thus far.

We present a method for identifying inaccurate data on the basis of qualitative correlations among related data. The method is based on the essential consideration that some data items within a dataset are qualitatively dependent: a set of data may describe the same phenomenon, or refer to the same behavior. For example, a patient's temperature, blood pressure and other symptomatic data reflect the patient's disease, and a couple of peaks on an infrared spectrum indicate the presence of a partial component. We call the dependency among data within a dataset *qualitative correlations among related data*[1]. By considering qualitative correlations among related data, we can obtain confirmatory or disconfirmatory evidence to identify inaccurate data. In general, related data should be simultaneously present or absent, so if most of the related data have been completely identified, these data will enhance the identification of the rest. For example, a benzene-ring can create many other peaks besides the strong peak in 3000 $cm^{-1}$- 3100 $cm^{-1}$. All the peaks created by the benzene-ring are related data which have qualitative correlations. If all the peaks except that in 3000 $cm^{-1}$- 3100 $cm^{-1}$ have been completely identified, the benzene-ring is quite likely to be contained by the unknown compound. Therefore, the inaccurate peak around 3000 $cm^{-1}$- 3100 $cm^{-1}$ may still be identified. In fact, spectroscopists frequently use the following knowledge in addition to the rules given at the beginning of this section:

> *If there is a strong peak around 3000 $cm^{-1}$- 3100 $cm^{-1}$, then the spectrum may be partially created by benzene-rings —– check peaks around 1650 $cm^{-1}$, 1550 $cm^{-1}$ and 700 $cm^{-1}$- 900 $cm^{-1}$ to make sure because a benzene-ring may have other peaks there at the same time.*

The central idea of our method is to find evidence for identifying inaccurate data by considering qualitative correlations among related data. The idea is very common in human thinking. When all the data except blood pressure of a patient show that the patient has a certain disease, we would naturally suspect that the blood pressure of the patient was inaccurately entered. Similarly, when all the peaks except one indicate that a partial component is present, we would naturally suspect that the unmatched peak was inaccurately measured or the peak was affected by noise or something else. If acceptable solutions can be made by assuming an inaccurate data item to be a reference value based on qualitative

1. Detailed definitions will be given later.

correlations between the data item and its related data, the inaccurate data item may be compensated and hence identified.

Our contributions include: (1) a method which assumes an inaccurate data item to be a certain reference value based on the qualitative correlations between the inaccurate data item and all of its related data, (2) an algorithm which crystallizes the method, and (3) a practical system which uses the algorithm to interpret infrared spectra.

The key point is a new concept called support coefficient function ($SCF$) for extracting, representing, and calculating qualitative correlations among related data. When measured data are inaccurate, the qualitative correlations among related data can provide evidence for confirming or disconfirming the hypothesis that the measured data are the same as the reference values. An approach to determining dynamic shift intervals of inaccurate data, an approach to calculating possibility of identifying inaccurate data, and an algorithm for identifying inaccurate data are proposed on the basis of $SCF$, respectively.

The method requires few assumptions in advance, so it can avoid inconsistency in knowledge and data bases. The method identifies inaccurate data by considering qualitative correlations among related data, so it is quite effective and efficient, especially in the case of problems where dependencies among data apparently exist. In general, qualitative correlations among data can always, more or less, be extracted. In the worst case where qualitative correlations are not known a priori, the method degenerates to a conventional fuzzy method[2].

We have developed a practical system for interpreting infrared spectra by using the method (Zhao & Nishida, 1994). The primary task of the system is to identify unknown compounds by interpreting their infrared spectra. We have fully tested the system against several hundred real spectra. The experimental results show that the method is significantly better than the traditional methods used in many similar systems. The rate of correctness ($RC$) and the rate of identification ($RI$) which are two important standards for evaluating the solutions of infrared spectrum interpretation are near 74% and 90% respectively, and the former is the highest among known systems.

In the following sections, we first describe the problem of identifying inaccurate data in Section 2. In Section 3 we give some definitions including the concept of support coefficient function ($SCF$) and other concepts based on $SCF$. In Section 4 we introduce our method for identifying inaccurate data by considering qualitative correlations among related data. Section 5 demonstrates the application of the method to a knowledge-based system for infrared spectrum identification, and shows the experimental results of the system. Related work is discussed in Section 6. Conclusions are addressed in Section 7.

## 2. Problem Description

In practical problems, measured data can be represented as a finite set:

2. We refer to the fuzzy methods which use an empirical fuzzy interval for each inaccurate data item as conventional fuzzy methods.

$$MD = \{d_1, d_2, ..., d_n\},$$

and reference values can also be represented as a finite set:

$$RV = \{r_1, r_2, ..., r_N\}.$$

Suppose interpreting or analyzing measured data is carried out on the basis of so-called "if-then" rules in which the premises are comparisons between $MD$ and $RV$ like "*if* $d_i = r_j$ *then* ...", or "*if* $(r_i \in MD) \wedge (r_j \in MD)$ *then* ...". When $MD$ is accurate, the main operation implied by these premises is usually to find a corresponding reference value from $RV$ for each data item in $MD$. However, when $MD$ is inaccurate, the operation becomes complicated. In this case, it is difficult to determine which reference value an inaccurate data item corresponds to, e.g., for some measured data no reference value may be simply identified, while for others more than one may be available.

For example, if received signals are known to be accurate, and an expected signal (reference value) can not be found from the signal series (measured data), then we can conclude that the expected signal does not appear. However, if received signals are inaccurate, and an expected signal can not be identified from the signal series, it is hard to decide whether the expected signal does not appear or appears but looks different due to the inaccuracy.

Most currently known approaches for dealing with inaccurate data such as fuzzy logic and probabilistic reasoning are mainly based on quantitative similarity or closeness between measured data and reference values. In some cases, however, the identity of qualitative features is more effective and reliable than quantitative similarity or closeness.

Consider signal analysis again. If an inaccurate signal has the same qualitative features as the expected one such as the interval of frequency, the signal may still be identified even though its quantitative features are slightly different from those of the expected one such as strength etc.; conversely, an inaccurate signal may not be identified if it is quantitatively similar to an expected signal but does not have the same qualitative features as the expected one.

We discussed the following points in Section 1, (1) some data items within a dataset are qualitatively dependent (i.e., they are related data), (2) there are qualitative correlations among related data, and (3) qualitative correlations among related data enable us to confirm or disconfirm the identity of qualitative features.

Therefore, $RV$ and $MD$ can be, explicitly or implicitly, divided into finite groups on the basis of qualitative dependencies among data, and the data in each group are related to each other. For example, $RV$ can be divided into $R_1$, $R_2$, ... and $R_k$:

$$RV = R_1 \cup R_2 \cup ... \cup R_k,$$

where

$$R_j = \{r_{j_l} \mid r_{j_l} \in RV, 1 \leq l \leq m\}.$$

The qualitative correlations among related data in $R_j$ include: (1) data in $R_j$ should be simultaneously present or absent which means that all reference values in $R_j$ should have corresponding data in $MD$, (2) the presence of $r_{j_p}$ may enhance the presence of $r_{j_q}$, and the absence of $r_{j_p}$ may depress the presence of $r_{j_q}$. Considering the qualitative correlations among related data will lead to evidence for the identification of inaccurate data.

The problem of interpreting/analyzing inaccurate data is to make qualitative hypotheses for $MD$, or in other words, to find a subset of $RV$ for $MD$, which is corresponding to $MD$:

$$IN(MD), \quad (IN(MD) \subset RV).$$

The problem can be briefly represented as the following predicate calculus:

$$\forall d_i \forall R_j ((d_i @ R_j) \wedge (R_j @ MD) \rightarrow R_j \subset IN(MD))^3,$$

where "$d_i@R_j$" and "$R_j@MD$" are two essential qualitative predicates in our method which represent that $d_i$ possibly (qualitatively) belongs to $R_j$ (i.e., ? $d_i \in R_j$), and $R_j$ possibly (qualitatively) belongs to $MD$ (i.e., ? $R_j \subset MD$), respectively. Determining "$A@B$" is based on qualitative correlations among related data. The work presented in this paper is mainly concentrated on determining "$d_i@R_j$" and "$R_j@MD$", and realizing the above predicate calculus.

## 3. Preliminaries

Before introducing our method, we first put forward and explain several new concepts in this section.

### 3.1 Qualitative Correlations among Related Data

**Definition 3.1** *Related data: If data $d_1$, $d_2$, ..., and $d_m$ describe a common phenomenon, or they refer to the same behavior simultaneously, then they can be treated as related data.*

For example, a patient's temperature, blood pressure and other symptomatic data are related data, and all the features for describing a criminal are also related data. The phenomenon that some data within a dataset are related data is more apparent in engineering. For instance, there are two types of related data in infrared spectrum interpretation as shown in Figure 1. First, as far as a single peak is concerned, the frequency (position) $f_i$, strength (height) $s_i$, and width (shape) $w_i$ of the peak are related data. Second, a partial component may create numerous peaks at the same time. If we consider all the peaks that a partial component may create, all of these peaks are related data.

**Definition 3.2** *Qualitative correlations among related data: If $d_i$ and $d_j$ are two related data items, then the presence of $d_i$ enhances the presence of $d_j$, and the absence of $d_i$ depresses the presence of $d_j$. This kind of effect is called qualitative correlations among related data.*

3. Conflicts (overlaps) in $IN(MD)$ should be eliminated. We will not discuss conflict-resolving in this paper, but will concentrate on the method for identifying inaccurate data, i.e., ? $d_i@R_j$ and ? $R_j@MD$. Interested readers may refer to the paper by Zhao (1994) for specific discussion concerning the problem of conflict resolution.

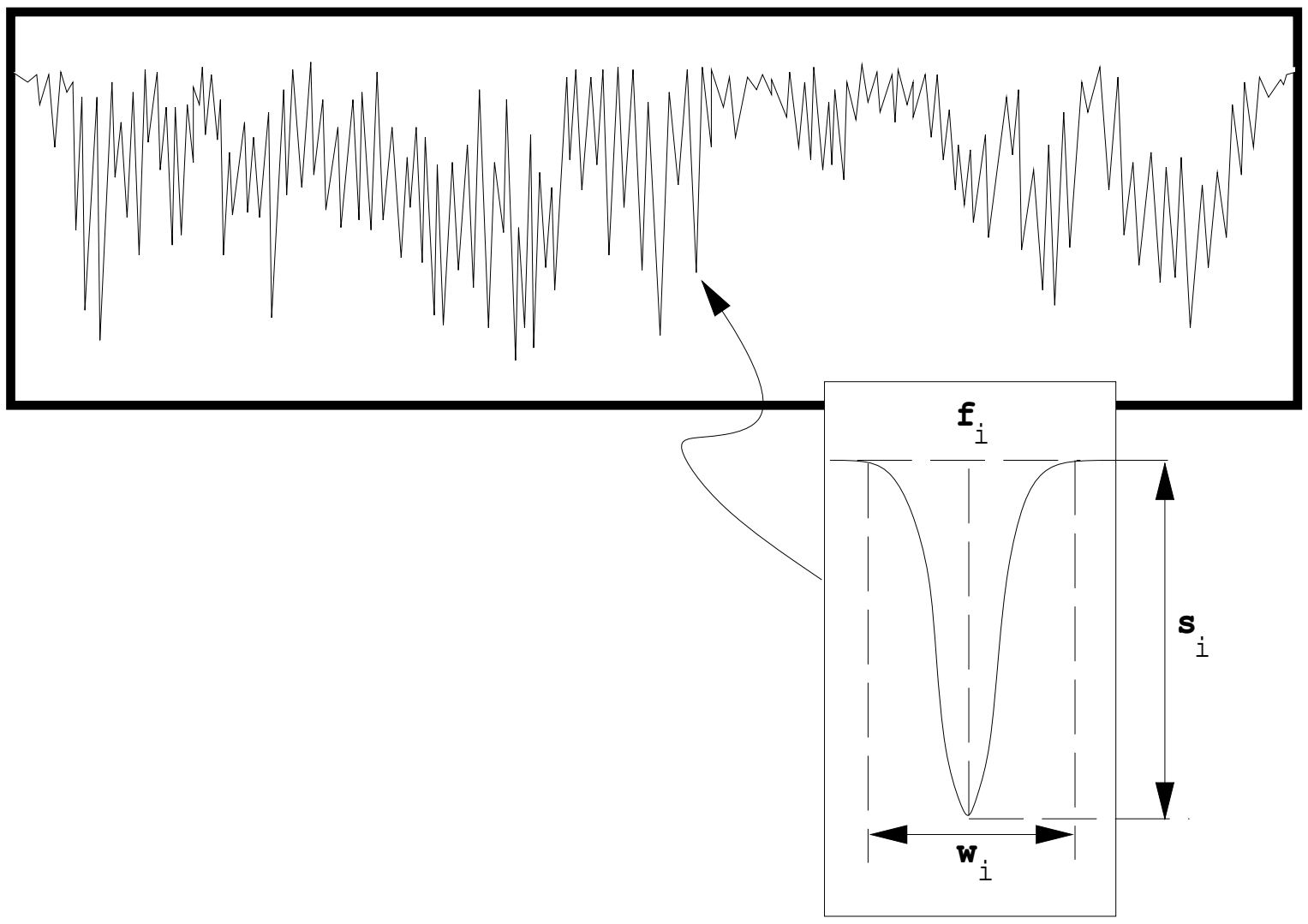

Figure 1: Example of related data in spectrum interpretation

Consider the above example of spectrum interpretation again. If spectral data are inaccurate (i.e., some measured peaks look like but are not exactly the same as reference peaks), considering qualitative correlations among related data may lead to qualitative evidence for the identification of inaccurate data. For example, suppose the frequency of a peak is slightly different from the reference value, and both the strength and width of the peak are the same as the reference values. Then the frequency of the peak may still be identified since both of its related data support it. Similarly, if peaks at low frequency sections are inaccurate, considering related peaks at high frequency sections may help identify these peaks, and vice versa.

### 3.2 Support Coefficient Function

**Definition 3.3** *Support coefficient function (SCF): If there are $m-1$ data related to $d_i$, then the support coefficient function of $d_i$ calculates the total effects from the related data by considering the qualitative correlations between $d_i$ and each of its related data.*

Suppose $\pi(d_i,\ d_j)$ represents the qualitative correlation between $d_i$ and $d_j$, then the support coefficient function of $d_i$ can be defined as:

$$SCF_i = \beta(\sum_{j=1, j\neq i}^{m} \pi(d_i, d_j), m).$$

$SCF_i$ should directly depend on how many and how much related data support $d_i$. When $SCF_i$ is greater than a certain value given by domain experts, the related data tend to support $d_i$; otherwise, the related data tend to depress $d_i$.

### 3.3 Evidence Based on SCF

In Section 2, we used "$d_i@R_j$" to express that $d_i$ can be qualitatively identified from $R_j$. Realizing "$d_i@R_j$" requires a definition of a shift interval $\triangle$ for $R_j$ such as:

$$R_j \pm \triangle = \{(r_{j_l} \pm \triangle) \mid l = 1, 2, ..., m\},$$

and a definition of the possibility of "$d_i \in R_j \pm \triangle$".

The above formula is similar to that in fuzzy logic, but contains completely different meanings. The primary difference is that the shift intervals are dynamically determined by $SCF_i$, while in fuzzy logic, the fuzzy intervals are usually provided by domain experts in advance or calculated with quantitative criteria.

**Definition 3.4** *Shift interval: Shift interval is a dynamic region for inaccurate data. Given a standard fuzzy interval for inaccurate data, the shift interval of $d_i$ varies around the standard fuzzy interval on the basis of $SCF_i$. When $SCF_i$ shows that the related data support $d_i$, the shift interval of $d_i$ becomes wider than the standard fuzzy interval. On the other hand, when $SCF_i$ shows that the related data do not support $d_i$, the shift interval of $d_i$ becomes narrower than the standard fuzzy interval.*

**Definition 3.5** *Evidence based on $SCF_i$: $SCF_i$ determines the shift interval of $d_i$, that is, $SCF_i$ determines how widely $d_i$ is allowed to shift. The wider the shift interval, the more easily $d_i$ is identified. Therefore, $SCF_i$ provides confirmatory or disconfirmatory evidence for identifying $d_i$.*

## 4. Making Qualitative Hypotheses for Inaccurate Data

In this section, we introduce and analyze our method for identifying inaccurate data. We first discuss the processes of realizing two essential predicates in our method, "$d_i@R_j$" and "$R_j@MD$" respectively. Then, we present an algorithm for making qualitative hypotheses for inaccurate data (i.e., for realizing the predicate calculus described in Section 2).

### 4.1 Predicate "$d_i@R_j$"

When $d_i$ is accurate, "$d_i@R_j$" is equal to "$d_i \in R_j$". If there is a reference value in $R_j$ which corresponds to $d_i$ (i.e., $r_{j_p} \in R_j$ and $r_{j_p} = d_i$), then $d_i@R_j = T$. If there is no reference value corresponding to $d_i$, then $d_i@R_j = F$. When $d_i$ is inaccurate, however, it is not sure whether $r_{j_p}$ corresponds to $d_i$. In this case, "$d_i@R_j$" means that $d_i$ possibly (qualitatively) belongs to $R_j$, or in other words, $r_{j_p}$ possibly (qualitatively) corresponds to $d_i$. The value of "$d_i@R_j$" is not $T$ or $F$, but the possibility of "$r_{j_p} = d_i$" or "$d_i \in R_j$".

We discussed in Section 2 that in some cases the identity of qualitative features is more robust and reliable than quantitative similarity or closeness. We have also discussed that qualitative correlations among related data can lead to evidence for the identity of qualitative features in diagnosis or interpretation. So if $r_{j_p}$ ($r_{j_p} \in R_j$) is assumed to correspond to $d_i$, and there are $m-1$ reference values ($r_{j_1}, r_{j_2}, ..., r_{j_{p-1}}, r_{j_{p+1}}, ..., r_{j_m}$) related to $r_{j_p}$, then each of the $m-1$ reference values should correspond to a certain data item in $MD$, and the $m-1$ data items in $MD$ are also related to each other. Therefore, qualitative correlations between $d_i$ and its $m-1$ related data items in $MD$ should be considered.

Our method first determines the possibility of "$r_{j_p} = d_i$" by calculating the similarity or closeness between $r_{j_p}$ and $d_i$ like conventional fuzzy methods, then considers qualitative correlations among related data to obtain evidence for updating the possibility. When the qualitative correlations show that the related data support "$r_{j_p} = d_i$", the possibility of "$r_{j_p} = d_i$" will increase. When the qualitative correlations show that the related data do not support "$r_{j_p} = d_i$", the possibility will decrease.

#### 4.1.1 Defining Support Coefficient Function

Suppose $r_{j_q}$ ($r_{j_q} \in R_j$) corresponds to $d_t$. Because $r_{j_q}$ is related to $r_{j_p}$, $d_t$ is related to $d_i$. As we have discussed, the qualitative correlation between $d_i$ and $d_t$ means that if $d_t$ exists, then $d_i$ is enhanced; otherwise, $d_i$ is depressed.

We first define the qualitative correlation between two related data items, $d_i$ and $d_t$, as:

$$c_i(d_t) = \begin{cases} 1 & \text{if } d_t \text{ can be found from } MD \text{ which satisfies: } r_{j_q} - d_o \leq d_t \leq r_{j_q} + d_o \\ 0 & \text{if } d_t \text{ can not be found from } MD \text{ which satisfies: } r_{j_q} - d_o \leq d_t \leq r_{j_q} + d_o \end{cases}$$

where $d_o$ is a standard fuzzy interval of inaccurate data, and $c_i(d_t)$ expresses the qualitative correlation between $d_i$ and $d_t$. $c_i(d_t)$=1 means that $d_i$ is enhanced since its related data item $d_t$ can be found from the measured dataset, and $c_i(d_t)$=0 means that $d_i$ is depressed since its related data item $d_t$ can not be found from the measured dataset. The definition of $c_i(d_t)$ is simply based on the consideration that if a data item is identified, then the data item will support its related data items (i.e., the coexisting data items).

As there are $m$ reference values in $R_j$, we can define the support coefficient function $SCF_i$ for $d_i$ based on $c_i(d_t)$ ($t = 1, 2, ..., m,\ t \neq i$):

$$SCF_i = \frac{1 + \sum_{t=1, t\neq i}^{m} c_i(d_t)}{m}$$

where $0 < SCF_i \leq 1$, and $SCF_i$ expresses the total qualitative correlations between $d_i$ and all of its related data. In other words, $SCF_i$ reflects the support coefficient of $r_{j_p}$ corresponding to $d_i$.

If $m = 1$, then $SCF_i = 1$. When $m > 1$, $SCF_i$ is in the direct ratio to the number of the related data which may be identified from $MD$.

#### 4.1.2 Determining Dynamic Shift Interval

Suppose $d_o$ is a standard fuzzy interval of inaccurate data, we define the dynamic shift interval of $d_i$ based on $SCF_i$ as:

$$\triangle d_i = \frac{(2m-1)d_o}{m} \times SCF_i$$

where $0 < \triangle d_i < 2d_o$, and $\triangle d_i$ is in the direct ratio to $SCF_i$.

If $m = 1$, then $SCF_i = 1$, and $\triangle d_i = d_o$. In other words, when qualitative correlations among data are not known a priori, $SCF_i = 1$ and $\triangle d_i = d_o$. In this case, the method degenerates to a conventional fuzzy method.

When $m$ is fixed, the more the related data are identified, the greater $SCF_i$ is, therefore the greater $\triangle d_i$ is. When $SCF_i$ is fixed, $\triangle d_i$ depends on the number of related data.

Table 1 shows the relation among $\triangle d_i$, $m$ and $SCF_i$.

| $\triangle d_i$ | | m | | | | | |
|---|---|---|---|---|---|---|---|
| | | 1 | 10 | 50 | 100 | 500 | 1000 |
| $SCF_i$ | 1 | $d_o$ | $1.9000d_o$ | $1.9800d_o$ | $1.9900d_o$ | $1.9980d_o$ | $1.9990d_o$ |
| | 0.8 | / | $1.5200d_o$ | $1.5840d_o$ | $1.5920d_o$ | $1.5984d_o$ | $1.5992d_o$ |
| | 0.5 | / | $0.9500d_o$ | $0.9900d_o$ | $0.9950d_o$ | $0.9990d_o$ | $0.9995d_o$ |
| | 0.3 | / | $0.5700d_o$ | $0.5940d_o$ | $0.5970d_o$ | $0.5994d_o$ | $0.5997d_o$ |
| | 0.1 | / | $0.1900d_o$ | $0.1980d_o$ | $0.1990d_o$ | $0.1998d_o$ | $0.1999d_o$ |

Table 1: Relation among $\triangle d_i$, $m$ and $SCF_i$

We can draw the following properties from the above formulas.

**Property 1**: *With the same $m$, the more the related data are identified, the greater $SCF_i$ is; otherwise, the smaller $SCF_i$ is.*

**Property 2**: *With the same $m$, the greater the $SCF_i$, the greater is $\triangle d_i$. In other words, the more the related data support $d_i$, the more widely $d_i$ is allowed to shift.*

**Property 3**: *With the same $SCF_i$, the greater the $m$, the less $\triangle d_i$ varies along with $m$. In other words, the greater the number of related data, the less a single related data item can affect $d_i$.*

Property 2 and Property 3 are illustrated in Figure 2.

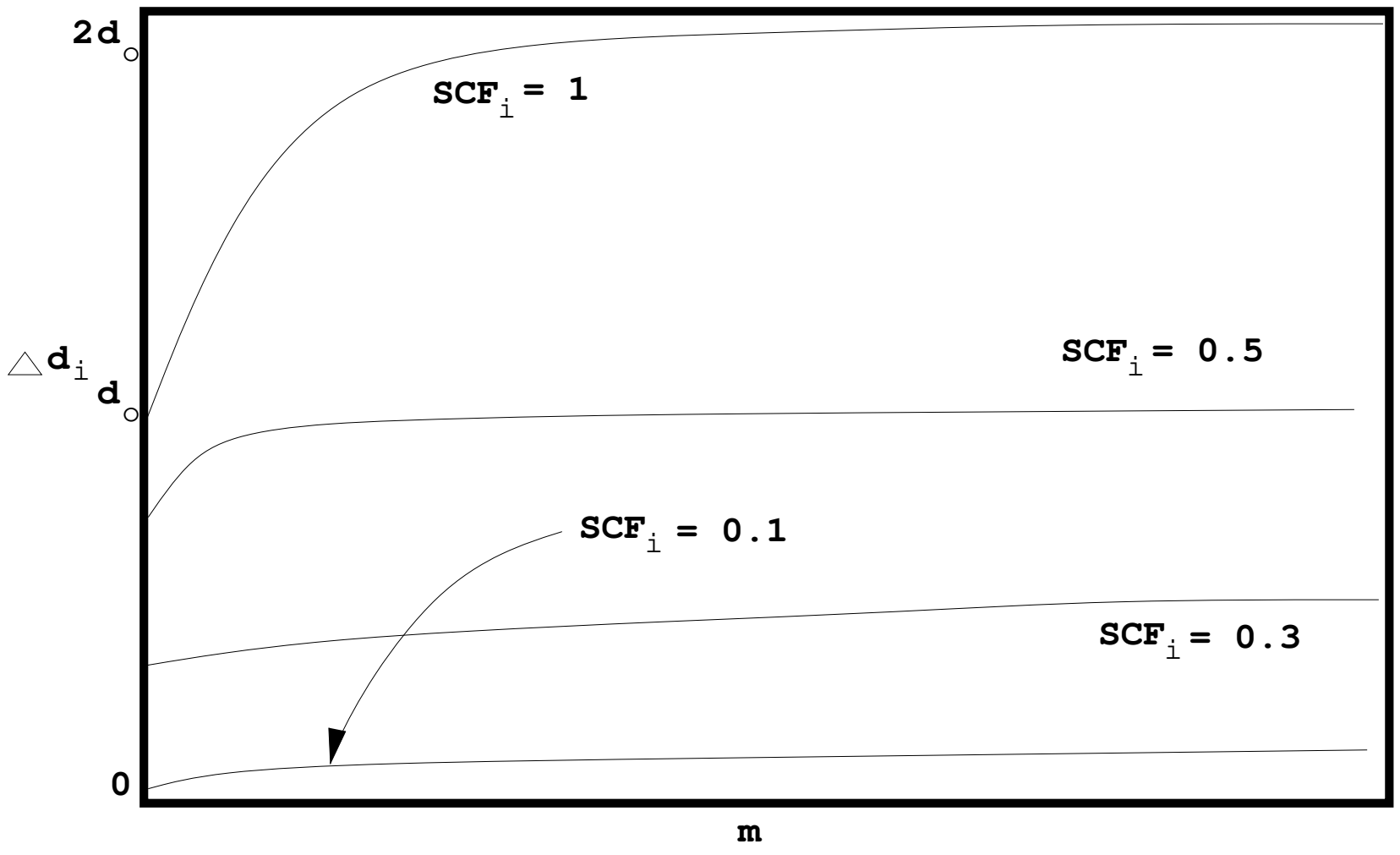

Figure 2: $\triangle d_i$ versus $m$ with different $SCF_i$

**Property 4**: *$\triangle d_i$ is in linear relation to $SCF_i$. The slope is equal to, or greater than 1.5, which means that $\triangle d_i$ heavily depends on $SCF_i$.*

**Property 5**: *Along with the increase of $m$, the slope increases very slightly. In other words, $\triangle d_i$ depends on the number of the related data which support $d_i$, rather than the total number of related data.*

Property 4 and Property 5 are illustrated in Figure 3.

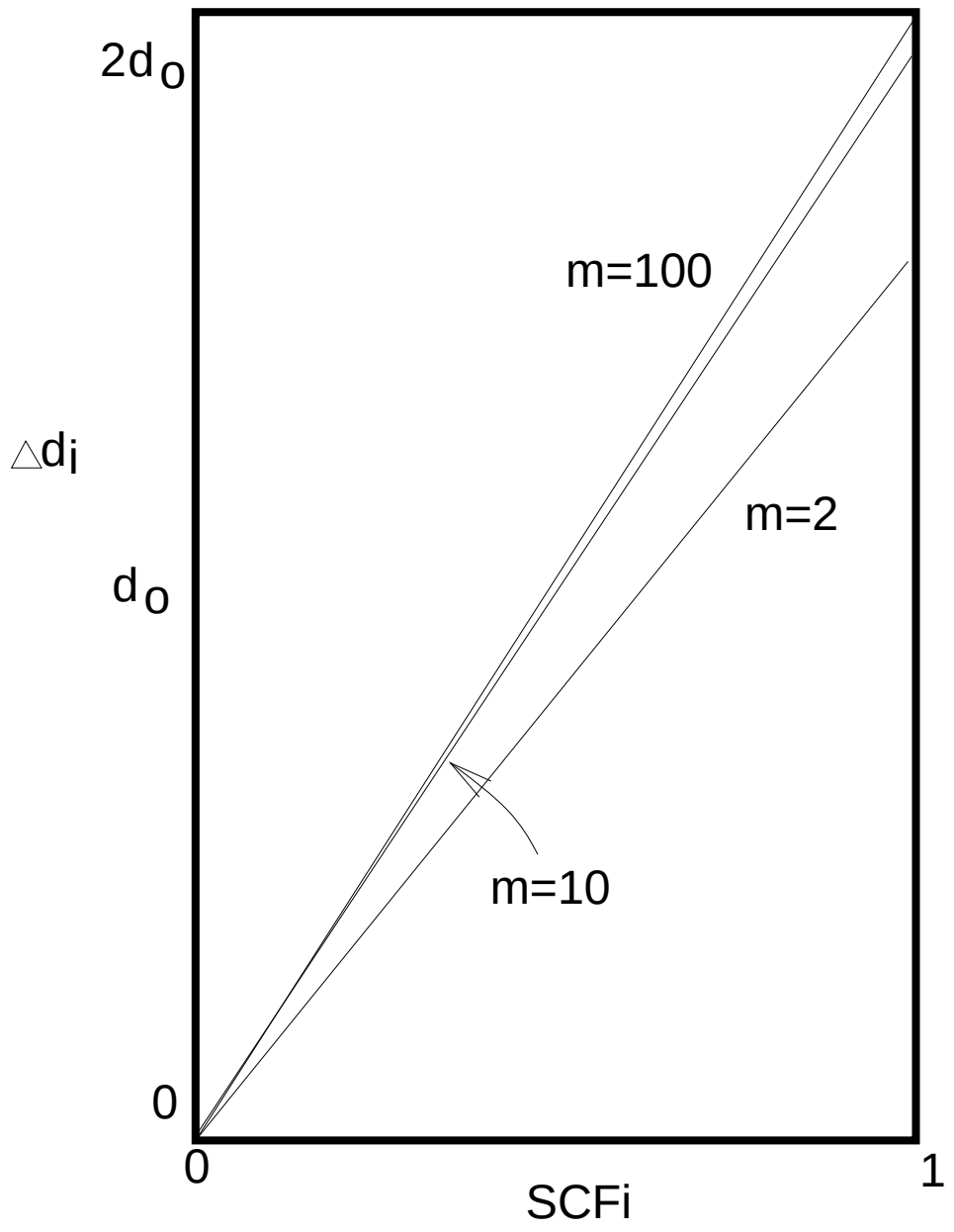

Figure 3: $\triangle d_i$ versus $SCF_i$ with different $m$

#### 4.1.3 Calculating Value of Predicate "$d_i@R_j$"

The value of "$d_i@R_j$" is equal to the possibility of "$r_{j_p} = d_i$" which can be calculated by using the following formula:

$$\mu_i = 1 - \frac{\mid d_i - r_{j_p} \mid}{\triangle d_i}$$

where $\mu_i \leq 1$.

At a glance, the representation of $\mu_i$ looks like the membership degree of "$r_{j_p} - \triangle d_i \leq d_i \leq r_{j_p} + \triangle d_i$" in fuzzy logic. However, the meaning is completely different, for $\triangle d_i$ is neither provided by domain experts nor determined by quantitative similarity or closeness. Here $\triangle d_i$ is determined on the basis of qualitative correlations among related data. When qualitative correlations among related data are not considered, $\triangle d_i$ is $d_o$, and the possibility is $1 - \frac{|d_i - r_{j_p}|}{d_o}$. With the consideration of qualitative correlations, the possibility is updated.

Two new properties can be drawn from the above formula for calculating $\mu_i$.

**Property 6**: *With the same $d_i$, the greater the $\triangle d_i$, the greater is $\mu_i$. In other words, the wider the dynamic shift interval, the greater is the value of "$d_i@R_j$". Formally, if $\triangle d_i'' \geq \triangle d_i' \geq \triangle d_i$, then $\mu_i'' \geq \mu_i' \geq \mu_i$.*

**Property 7**: *$SCF_i$ provides qualitative evidence for accepting or rejecting $d_i$ as $r_{j_p}$ since $\mu_i$ is in the direct ratio to $\triangle d_i$, and $\triangle d_i$ is in the direct ratio to $SCF_i$.*

Property 6 and Property 7 are illustrated in Figure 4.

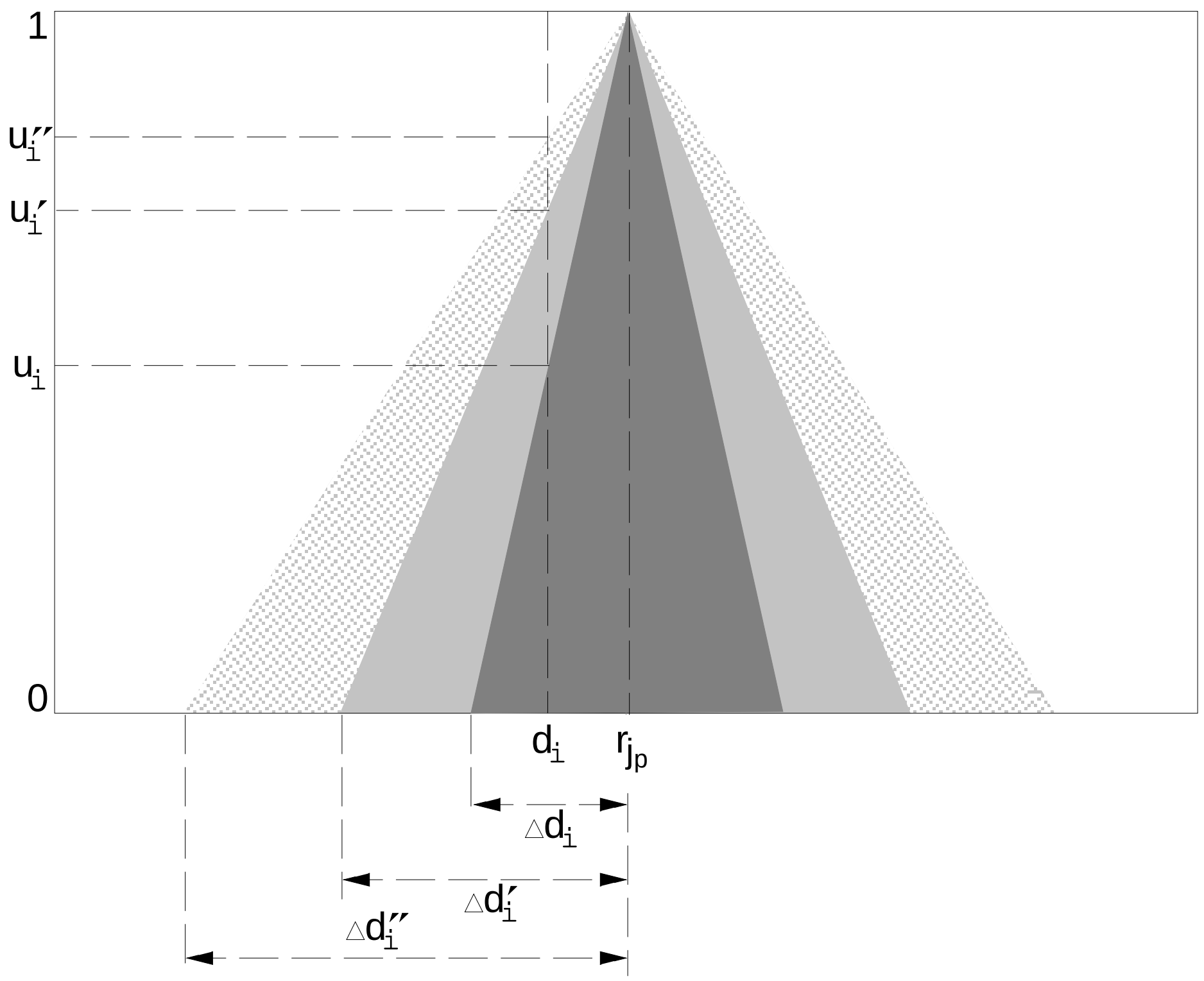

Figure 4: Value of "$d_i@R_j$" versus various $\triangle d_i$

The above process of realizing "$d_i@R_j$" and calculating the value of "$d_i@R_j$" can be expressed by the following procedure.

**Procedure** $d_i@R_j$

$select\ r_{j_p}\ from\ R_j;$
$SCF_i = 0;$
$if\ d_i = r_{j_p}\{$
$\quad SCF_i = 1;$
$\quad \mu_i = 1;$
$\}$
$else\{$
$\quad for\ each\ r_{j_l} \in R_j\ (l = 1, ..., m,\ l \neq p)\{$
$\quad\quad calculate\ c_i(d_t)$[4]$;$
$\quad\quad SCF_i = SCF_i + c_i(d_t);$
$\quad \}$
$\quad SCF_i = (1 + SCF_i)/m;$
$\quad \triangle d_i = d_o \times SCF_i \times (2m-1)/m;$
$\quad \mu_i = 1 - \mid d_i - r_{j_p} \mid / \triangle d_i;$
$\}$

4. $d_t$ stands for the data item in $MD$ which corresponds to $r_{j_l}$.

$if\ \mu_i > 0$

$return\ \mu_i;$

$else$

$return\ NIL$

**end procedure**

When $d_i$ can be identified with a certain possibility (i.e., $\mu_i > 0$), the procedure returns $T$ (i.e., the value of $\mu_i$); otherwise, the procedure returns $F$.

### 4.2 Predicate "$R_j@MD$"

When $MD$ is accurate, "$R_j@MD$" is equal to "$R_j \subset MD$". If all the $m$ reference values in $R_j$ can be identified from $MD$, then $R_j@MD = T$; otherwise $R_j@MD = F$. When $MD$ is inaccurate, however, "$R_j@MD$" means that $R_j$ is possibly (qualitatively) a subset of $MD$. The value of "$R_j@MD$" is not $T$ or $F$, but the possibility that all the reference values in $R_j$ can be identified from $MD$.

If $\mu_l > 0$ ($l = 1, 2, ..., m$), then $R_j$ can be regarded as a subset of $MD$ with a certain possibility. Let $s_1$, $s_2$, ..., and $s_m$ be the priorities of the reference values in $R_j$, then the value of "$R_j@MD$" can be calculated based on $\mu_1$, $\mu_2$, ..., and $\mu_m$ by using the following formula:

$$R_j@MD = \frac{\sum_{l=1}^{m} s_l \times \mu_l}{\sum_{l=1}^{m} s_l}, \qquad s_l > 0, \quad \mu_l > 0.$$

Suppose $\mu_i$ has been calculated by using **procedure** $d_i@R_j$, then the process of realizing "$R_j@MD$" and calculating the value of "$R_j@MD$" can be expressed by a simple procedure.

**Procedure** $R_j@MD$

$P = s_i \times \mu_i;$

$S = s_i;$

$for\ l = 1\ to\ m\ (l \neq p)\{$

$\mu_l = d_t@R_j;$

$if\ \mu_l > 0\{$

$P = P + s_l \times \mu_l;$

$S = S + s_l;$

$\}$

$else\{$

$P = 0;$

$exit;$

$\}$

$\}$

$if\ P > 0$

$return\ P/S;$

$else$

$return\ NIL$

**end procedure**

When $R_j$ can be identified as a subset of $MD$ with a certain possibility (i.e., $P/S$), the procedure returns $T$ (i.e., the value of $P/S$); otherwise, the procedure returns $F$.

### 4.3 Algorithm for Making Qualitative Hypotheses for Inaccurate Data

We give the following algorithm for interpreting/analyzing measured data based on **procedure** $d_i@R_j$ and **procedure** $R_j@MD$. When measured data are not accurate, the algorithm can identify inaccurate data items by considering qualitative correlations among related data.

**Algorithm** *Making-Qualitative-Hypotheses*

$IN(MD) = \emptyset$;

*for* $i = 1$ to $n$ {

  *for* $j = 1$ to $k$ {

    $P(R_j) = 0$;

    *if* $d_i@R_j$ (*i.e.*, **Procedure** $d_i@R_j$)

      *if* $R_j@MD$ (*i.e.*, **Procedure** $R_j@MD$) {

        $R_j \rightarrow IN(MD)$;

        $P(R_j) = R_j@MD$;

      }

      *end if*

    *end if*

  }

  *end for*

}

*end for*

**end algorithm**

In the algorithm, $P(R_j)$ represents the value of "$R_j@MD$". The algorithm is actually the realization of the predicate calculus: $\forall d_i \forall R_j((d_i@R_j) \wedge (R_j@MD) \rightarrow R_j \subset IN(MD))$.

For each measured data item in $\{d_1, d_2, ..., d_n\}$, the algorithm searches $\{R_1, R_2, ..., R_k\}$ once. For each $R_j$ ($R_j = \{r_{j_1}, r_{j_2}, ..., r_{j_m}\}$), the algorithm checks other $n-1$ measured data items for $m$ times, and other $m-1$ reference values for $n$ times. Therefore, with blind search, the number of operations is about (at worst): $n \times k \times [m \times (n-1) + n \times (m-1)] = 2 \times k \times m \times n^2 - k \times n^2 - k \times m \times n$. Since $k$ and $m$ are two constants, the complexity of the algorithm is $O(n^2)$.

## 5. Application to Infrared Spectrum Interpretation

We have developed a knowledge-based system for interpreting infrared spectra by applying the proposed method, and have fully tested the system against several hundred real spectra. The experimental results show that the proposed method is significantly better than the conventional methods used in many similar systems.

### 5.1 Infrared Spectrum Interpretation

The primary task of infrared spectrum interpretation is to identify unknown objects by interpreting their infrared spectra. In this paper, we will limit the problem to interpretation of infrared spectra of compounds to determine composition of unknown compounds without loss of generality.

Selecting infrared spectrum interpretation as the domain of application is out of the following reasons:

1. Interpreting infrared spectra is a very significant problem in both academic research and industrial application. For example, in chemical science and engineering, interpreting infrared spectra of compounds is the most effective way to identify unknown compounds, and to analyze the composition and purity of compounds (Colthup, Daly, & Wiberley, 1990).

2. Interpreting infrared spectra is a very difficult problem. First, spectral data are huge in quantity, and complex in representation. Second, both symbolic reasoning and numerical analysis are needed to interpret infrared spectral data (Puskar, Levine, & Lowry, 1986; Sadtler, 1988).

3. Interpreting infrared spectra is a typical problem dealing with inaccurate data since spectral data are often inaccurate. They often shift from their theoretical values due to various reasons. For example, the following is an assertion for spectrum interpretation:

   *The high frequency peak of partial component $PC_\alpha$ is located at $F_i$.*

   In practice, however, the peak of $PC_\alpha$ may irregularly shift around $F_i$ due to noise or other unforeseen reasons. When the above assertion is used to identify real spectra, uncertainty arises.

### 5.2 Applying the Proposed Method to Infrared Spectrum Interpretation

Interpreting infrared spectra is a special problem of diagnosis. Suppose the infrared spectrum of an unknown compound can be thresholded and represented as a finite set of peaks (i.e., the measured dataset $MD$):

$$Sp = \{p_1, p_2, ..., p_n\},$$

where every peak consists of the frequency (position) $f$, strength (height) $s$, and width (shape) $w$, respectively:

$$p_i = (f_i, s_i, w_i) \qquad i = 1, 2, ..., n.$$

Because $f_i$, $s_i$ and $w_i$ refer to the same peak $p_i$, they are related data. This is the first kind of related data in infrared spectrum interpretation.

Suppose there are finite partial components (i.e., reference values $RV$):

$$\begin{aligned} PC &= \{PC_1, PC_2, ..., PC_k\} \\ &= \{\{p_{j_1}, p_{j_2}, ..., p_{j_m}\} \mid j = 1, 2, ..., k\} \\ &= \{\{(f_{j_p}, s_{j_p}, w_{j_p}) \mid p = 1, 2, ..., m\} \mid j = 1, 2, ..., k\}. \end{aligned}$$

Because $f_{j_p}$, $s_{j_p}$ and $w_{j_p}$ also refer to the same reference peak $p_{j_p}$, they are the first kind of related data as well.

The spectroscopic knowledge for interpreting infrared spectra is usually expressed as "*if $p_i$ is equal to $p_{j_p}$, then $p_i$ may be created by partial component $PC_j$*". Here "$p_i$ is equal to $p_{j_p}$" represents that $f_i$, $s_i$, and $w_i$ are equal to $f_{j_p}$, $s_{j_p}$, and $w_{j_p}$ respectively.

The first kind of related data has the following qualitative correlations:

1. $f_i$, $s_i$ and $w_i$ should be identified simultaneously, that is,
   - if $f_i$ is $f_{j_p}$, then $s_i$ is $s_{j_p}$ and $w_i$ is $w_{j_p}$, and
   - if $s_i$ is $s_{j_p}$, then $f_i$ is $f_{j_p}$ and $w_i$ is $w_{j_p}$, and
   - if $w_i$ is $w_{j_p}$, then $f_i$ is $f_{j_p}$ and $s_i$ is $s_{j_p}$.
2. related data support each other. For example, if both $f_i$ and $s_i$ have been identified, then they will enhance the identification of $w_i$. Conversely, if $f_i$ and $s_i$ have not been identified, then they will weaken the identification of $w_i$.

Our method for identifying $f_i$, $s_i$ and $w_i$ based on the qualitative correlations among them can be formalized as the following predicate calculi, respectively:

$$\begin{aligned} &\forall f_i \forall p_{j_p}((f_i @ p_{j_p}) \wedge (p_{j_p} @ p_i) \rightarrow p_i \textit{ is created by } PC_j), \text{ and} \\ &\forall s_i \forall p_{j_p}((s_i @ p_{j_p}) \wedge (p_{j_p} @ p_i) \rightarrow p_i \textit{ is created by } PC_j), \text{ and} \\ &\forall w_i \forall p_{j_p}((w_i @ p_{j_p}) \wedge (p_{j_p} @ p_i) \rightarrow p_i \textit{ is created by } PC_j), \end{aligned}$$

where "*$p_i$ is created by $PC_j$*" means that $f_i$, $s_i$ and $w_i$ can be qualitatively identified to be $f_{j_p}$, $s_{j_p}$ and $w_{j_p}$.

In general, each partial component may create finite peaks at the same time. So if $p_i$ is created by $PC_j$, then $Sp$ is partially created by $PC_j$; if $Sp$ is partially created by $PC_j$, then all the peaks that $PC_j$ may create should be contained by $Sp$ simultaneously. Therefore, all the peaks created by a partial component are also related data. This is the second kind of related data in infrared spectrum interpretation.

The second kind of related data has the following qualitative correlations:

1. all the peaks of a partial component should be identified simultaneously, that is,

   if $p_i$ is $p_{j_p}$, then $p_{j_l} \in Sp$ $(l = 1, 2, ..., m, l \neq p)$.

2. the peaks created by the same partial component support each other. For example, if most of the peaks of a partial component have been identified, these peaks will enhance the identification of the rest peaks. Conversely, if most of the peaks of a partial component can not be identified, then the identification of the rest peaks will be depressed.

Our method for identifying related peaks based on the qualitative correlations can be formalized as the following predicate calculus:

$$\forall p_i \forall PC_j((p_i @ PC_j) \wedge (PC_j @ Sp) \rightarrow PC_j \subset IN(Sp)).$$

### 5.3 System for Interpreting Infrared Spectra

Our system is implemented with C and MS-WINDOWS. Figure 5 shows the data flow diagram of the system.

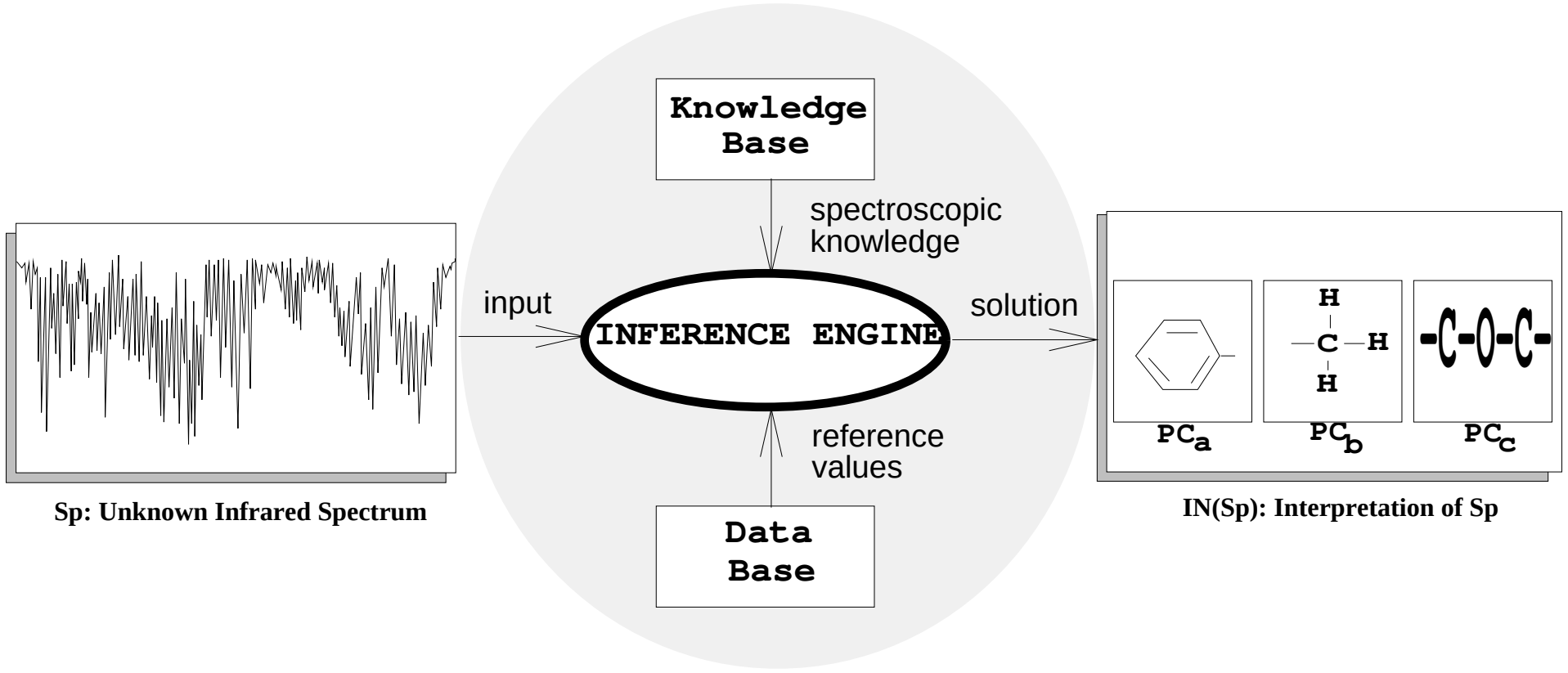

Figure 5: Data flow diagram of the system

The input data of the system are infrared spectra of unknown compounds, and the solutions are partial components that the input spectra may contain. Because inferences are based on qualitative features of spectral data and qualitative correlations among related data, the system can gain high correct interpretation performance with noisy spectral data.

As we mentioned before, there are two types of related data in infrared spectrum interpretation: all the features of a single peak (i.e., $f_i$, $s_i$ and $w_i$ of $p_i$), and all the peaks of a single partial component (i.e., $p_1$, $p_2$, ... and $p_m$). The inference engine of the system employs the proposed method to both types of the related data when inaccuracy arises.

### 5.4 An Example

We discuss the performance of the system through the following example. Figure 6 shows an infrared spectrum of an unknown compound. The spectrum is very hard to interpret since the peak with an arrow (named $p_1$) shifts substantially. Our system correctly identifies that $p_1$ is created by partial component benzene-ring.

In contrast, many similar systems can not correctly identify the peak (Clerc, Pretsch, & Zurcher, 1986; Hasenoehrl, Perkins, & Griffiths, 1992; Wythoff, Buck, & Tomellini, 1989) since the peak of a benzene-ring at this frequency position (named $p_{b_1}$) should be a strong peak (i.e., $s_{b1} > 1.000$) according to spectroscopic knowledge, not a medium one ($s_1 = 0.510$) as the case in this example. Systems based on conventional fuzzy methods usually assume a fuzzy interval for each inaccurate peak, then determine the membership degree that the inaccurate peak is in the fuzzy interval. Suppose the reference value for a strong peak is 1.000, and the fuzzy interval for a strong peak is 0.300 (Colthup, Daly,

& Wiberley, 1990), then only peaks with strength of $1 \pm 0.300$ can be regarded as strong peaks. Obviously, by conventional fuzzy methods, the possibility of $p_1$ being a strong peak is zero, i.e., $\mu_{benzene-ring}(s_1) = 0$.

Inferring on the basis of qualitative correlations among related data, our system makes a correct interpretation of the spectrum. Through the following two cases, we introduce the inference process of the system, and at the same time demonstrate the use of our method for identifying inaccurate data.

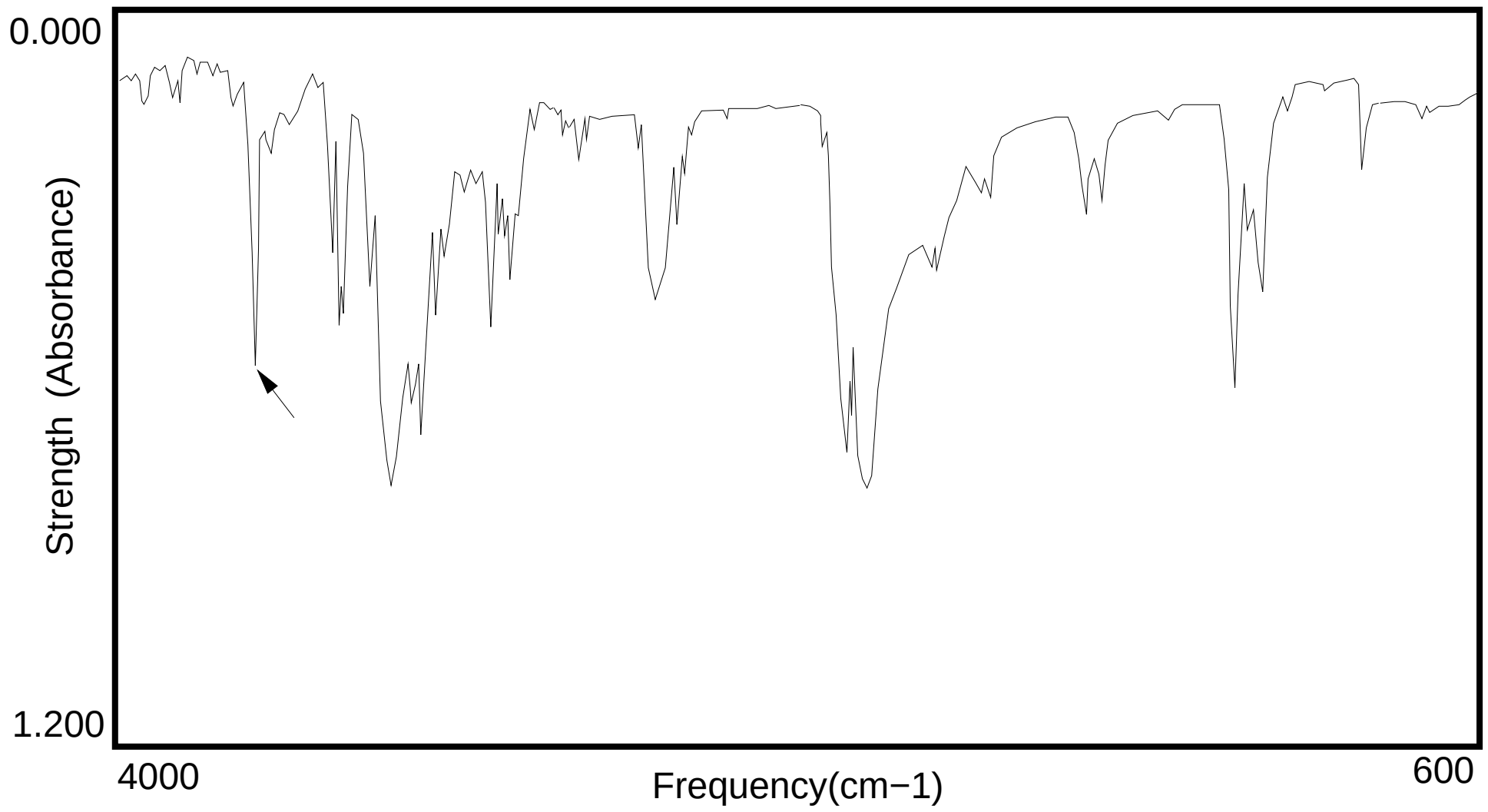

Figure 6: An example of infrared spectrum

### 5.4.1 Case I: Considering the First Kind of Related Data

Because the frequency (position) and width (shape) of $p_1$ are both the same as those of benzene-ring, the possibility of $f_1$ being identified as $f_{b1}$ is 100% (i.e., $\mu_{benzene-ring}(f_1) = 1$), and the possibility of $w_1$ being identified as $w_{b1}$ is also 100% (i.e., $\mu_{benzene-ring}(w_1) = 1$[5].

As we have discussed before, $f_1$, $s_1$ and $w_1$ are related data, so we can obtain confirm evidence for identifying $s_1$ by considering qualitative correlations among $s_1$, $f_1$ and $w_1$:

$\mu_{benzene-ring}(f_1) = 1$,

so, $c_{s_1}(f_1) = 1$ ($c_{s_1}(f_1)$ represents the qualitative correlation between $s_1$ and $f_1$),

$\mu_{benzene-ring}(w_1) = 1$,

so, $c_{s_1}(w_1) = 1$ ($c_{s_1}(w_1)$ represents the qualitative correlation between $s_1$ and $w_1$)

so, $SCF_{s_1} = \frac{1+2}{3} = 1$, and

$\triangle s_1 = \frac{(6-1)\times 0.300}{3} \times 1 = 0.500$, and

$s_1@p_{b_1} = 1 - \frac{1-0.510}{0.500} = 0.02$.

5. $\mu_*(d)$ means the possibility of $d$ being identified by conventional fuzzy methods, i.e., $SCF$ is not considered.

By considering $SCF_{s_1}$, the possibility of $p_1$ being regarded as a strong peak of benzene-ring increases from 0 to 0.02. As possibility, 0.02 may not be different from 0.04 or 0.06, but 0.02 is significantly different from 0. Many near-misses may be handled by the negligible possibility. For example, in most systems based on fuzzy and other methods (Clerc, Pretsch, & Zurcher, 1986), it is impossible to identify $p_1$ to be "strong" (i.e., $\mu_{benzene-ring}(s_1) = 0$), but considering qualitative correlations among related data makes it possible although the possibility is only 0.02.

As mentioned before, $f_1$ and $w_1$ are both the same as the reference values, so $f_1@p_{b_1} = 1$, and $w_1@p_{b_1} = 1$.

Suppose the priorities of $f_1$, $s_1$ and $w_1$ are 2, 1 and 1 respectively, then the possibility of $p_1$ being identified as $p_{b_1}$ is:

$$\mu_1 = p_{b_1}@p_1 = \frac{2 \times 1 + 0.02 + 1}{4} = 0.755.$$

#### 5.4.2 Case II: Considering the Second Kind of Related Data

The process of considering the second kind of related data is quite similar.

We have got that the possibility of $p_1$ being created by a benzene-ring is $\mu_1$ ($\mu_1 = 0.755$). Suppose the benzene-ring can create $m$ peaks: $\{p_{b_1}, p_{b_2}, ..., p_{b_m}\}$, then the $m$ peaks are related to each other. If $p_1$ is created by the benzene-ring, then $Sp$ is partially created by the benzene-ring, i.e., the benzene-ring is contained by the unknown spectrum; if $Sp$ is partially created by the benzene-ring, then the other $m - 1$ peaks of the benzene-ring should also be identified.

By using the same procedure as obtaining $\mu_1$, we can get $\mu_2$, $\mu_3$, ... and $\mu_m$ as well. According to our method, the qualitative correlation between two related peaks, $p_i$ and $p_j$, is defined as:

$$c_i(p_j) = \begin{cases} 1 & if \ \ \mu_j \geq 0.5 \\ 0 & if \ \ \mu_j < 0.5. \end{cases}$$

So

$$SCF_i = \frac{1 + \sum_{j=1, j \neq i}^{m} c_i(p_j)}{m}, \quad 0 < SCF_i \leq 1.$$

Let $d_o = 1$, then

$$\triangle d_i = \frac{2m - 1}{m} \times SCF_i, \quad 0 < \triangle d_i < 2,$$

and

$$p_i@benzene - ring = 1 - \frac{1 - \mu_i}{\triangle d_i}, \quad p_i@benzene - ring \leq 1.$$

Roughly, when $SCF_i > 0.5$, related peaks tend to support $p_i$. When related peaks support $p_i$, $\triangle d_i > 1$. When $\triangle d_i > 1$, $p_i@benzene - ring > \mu_i$.

Table 2 shows the relation among $p_i@benzene - ring$, $\mu_i$ and $\triangle d_i$.

| $p_i@benzene - ring$ | | $\mu_i$ | | | | |
|---|---|---|---|---|---|---|
| | | 1 | 0.8 | 0.5 | 0.3 | 0 |
| $\triangle d_i$ | 1.3 | 1 | 0.846 | 0.615 | 0.462 | 0.231 |
| | 1.1 | 1 | 0.818 | 0.545 | 0.364 | 0.091 |
| | 1 | 1 | 0.8 | 0.5 | 0.3 | 0 |
| | 0.9 | 1 | 0.778 | 0.444 | 0.222 | -0.111 |
| | 0.7 | 1 | 0.714 | 0.286 | 0 | -0.429 |

Table 2: Relation among $p_i@benzene - ring$, $\mu_i$ and $\triangle d_i$

In the above example, $SCF_1 = 0.850$, and $\triangle d_1 = 1.658$, so

$$p_1@benzene - ring = 1 - \frac{1 - 0.755}{1.658} = 0.852.$$

Therefore, the possibility of $p_1$ being identified as $p_{b_1}$ increases from 0.755 to 0.852 due to qualitative correlations among related peaks. The process is similar to the probability propagation in probabilistic reasoning. Here identifying $p_1$ is a hypothesis, and qualitative correlations among related data of $p_1$ are pieces of evidence.

After all the peaks of the benzene-ring are identified, the possibility that the benzene-ring is contained by $Sp$ can be finally calculated by employing the same method as described in Section 5.4.1.

### 5.5 Analysis of Experimental Results

We compare two methods in the experiments. The first method (called "$AF$") is a conventional fuzzy method which is used by most similar systems (Clerc, Pretsch, & Zurcher, 1986; Wythoff, Buck, & Tomellini, 1989). To use $AF$, each reference value must be associated with a fuzzy interval for dealing with inaccuracy. Both reference values and fuzzy intervals are empirically determined (Colthup, Daly, & Wiberley, 1990).

Table 3 lists some reference values and their fuzzy intervals used by $AF$.

| | | | |
|---|---|---|---|
| $CH_3$ | $2960 \pm 15cm^{-1}$ | $strong \pm 0.3$ | $sharp \pm 1$ |
| | $2870 \pm 15cm^{-1}$ | $strong \pm 0.3$ | $sharp \pm 1$ |
| | $1450 \pm 10cm^{-1}$ | $medium \pm 0.3$ | $sharp \pm 0.5$ |
| | ... | | |
| $benzene - ring$ | $3055 \pm 25cm^{-1}$ | $strong \pm 0.3$ | $sharp \pm 1.5$ |
| | $1645 \pm 10cm^{-1}$ | $medium \pm 0.3$ | $sharp \pm 0.5$ |
| | $1550 \pm 30cm^{-1}$ | $medium \pm 0.3$ | $sharp \pm 1$ |
| | $1450 \pm 3cm^{-1}$ | $medium \pm 0.3$ | $sharp \pm 0$ |
| | ... | | |
| $-CH_2 - OH$ | $3635 \pm 5cm^{-1}$ | $strong \pm 0.3$ | $broad \pm 1$ |
| | $3550 \pm 25cm^{-1}$ | $strong \pm 0.3$ | $sharp \pm 1$ |
| | ... | | |

Table 3: Some reference values and their fuzzy intervals

The membership function of $AF$ is:

$$\mu_r(d) = max\{0, 1 - \frac{\mid d - r \mid}{\triangle d}\},$$

where $d$ is a measured data item, $r$ is a reference value, $\triangle d$ is the fuzzy interval of $r$, and $0 \leq \mu_r(d) \leq 1$.

The second method (called "$AF^*$") is the proposed method. $AF^*$ uses the same reference values and fuzzy intervals as $AF$, but the fuzzy intervals in $AF^*$ are only used as standard fuzzy intervals based on which dynamic shift intervals are determined by considering qualitative correlations among related data.

$AF$ and $AF^*$ use the same reference values and empirical fuzzy intervals. The formula for calculating membership degrees in $AF$ (i.e., $\mu_r(d) = max\{0, 1 - \frac{|d-r|}{\triangle d}\}$) is also similar to the formula for calculating possibility in $AF^*$ (i.e., $\mu_i = 1 - \frac{|d_i - r_{j_p}|}{\triangle d_i}$). However, in $AF$, $\triangle d$ is simply an empirical fuzzy interval, while in $AF^*$, $\triangle d_i$ is a dynamic shift interval based on qualitative correlations among related data.

We have tested the system against several hundred real infrared spectra of organic compounds. The experimental results show that $AF^*$ is significantly better than $AF$.

Table 4 lists part of the experimental results in which the first column indicates the solutions obtained by $AF$; the second column indicates the solutions obtained by $AF^*$; and the third column shows the correct solutions.

| *AF* (Without *SCF*) | *AF** (With *SCF*) | Correct Solutions |
|---|---|---|
| ● -CH2- CH3- -[CH2]n- | ● -CH2- CH3- -[CH2]n- | -CH2- CH3- -[CH2]n- |
| (2/3) -CH2- -C- | ● -CH2- CH3- -C- | -CH2- CH3- -C- |
| ● -CH2- CH3- CH3<br>-CH<br>CH3 | ● -CH2- CH3- CH3<br>-CH<br>CH3 | -CH2- CH3- CH3<br>-CH<br>CH3 |
| ● -CH2- CH3- -C- | ● -CH2- CH3- -C- | -CH2- CH3- -C- |
| (2/3) CH3- CH3<br>-CH<br>CH3 | ● CH3- CH3<br>-CH<br>CH3 ⌬ | CH3- CH3<br>-CH<br>CH3 ⌬ |
| (3/4) -CH2- CH3- >C=CH- | ● -CH2- CH3- >C=CH- ⌬ | -CH2- CH3- >C=CH- ⌬ |
| ● CH3- ⌬ | (2/2) -CH2- CH3- ⌬ | CH3- ⌬ |
| (2/3) -CH2- CH3- | (2/3) -CH2- CH3- | -CH2- CH3- ⌬ |
| (1/3) >C=CH- | (2/3) CH3- >C=CH- | CH3- >C=CH- ⌬ |
| ● -[CH2]n- -C=CH ⌬ | ● -[CH2]n- -C=CH ⌬ | -[CH2]n- -C=CH ⌬ |
| (4/5) -CH2- CH3- >C=CH- -CH[CH3]2 | ● -CH2- CH3- >C=CH- -CH[CH3]2 ⌬ | -CH2- CH3- >C=CH- -CH[CH3]2 ⌬ |
| (2/3) -CH2- ⌬⌬ | ● -CH2- CH3- ⌬⌬ | -CH2- CH3- ⌬⌬ |
| (1/2) C(Cl)(Cl) | ● -C=C- C(Cl)(Cl) | -C=C- C(Cl)(Cl) |
| (3/4) CH3- ⌬ NH2- | ● CH3- NH2- -C- ⌬ | CH3- NH2- -C- ⌬ |

● : identified PC set is the same as the PC set in the correct solution(in this case, RI=1)

(n) : identified PC set is not the same as the PC set in the correct solution(the number indicates the RI)

Table 4: Experimental results with $AF$ and $AF^*$

There are two important standard metrics for evaluating solutions of infrared spectrum interpretation:

**Definition 5.1** *Rate of correctness (RC): the rate that the identified partial component set is exactly the same as the partial component set in the correct solutions.*

**Definition 5.2** *Rate of identification (RI): the rate that the partial components in the correct solutions are identified.*

Table 5 shows the comparison between $AF$ and $AF^*$ with the two standard metrics.

| | RC | (error-rate) | RI | (error-rate) |
|---|---|---|---|---|
| $AF$ | 0.455 | (0.545) | 0.812 | (0.188) |
| $AF^*$ | 0.736 | (0.264) | 0.894 | (0.106) |

Table 5: Evaluation of $AF$ & $AF^*$ with $RC$ and $RI$

Table 5 demonstrates that both the $RC$ and $RI$ increase by integrating $SCF$, but the $RC$ increases more significantly. The reason is that although $AF$ can identify most partial components of unknown compounds, the rate that it can identify all partial components of unknown compounds is low because there are always some partial components whose measured peaks seriously shift from the reference values.

### 5.6 Comparison with Related Systems

Related systems mainly fall into the following four categories: (1) Systems based on Y/N classification, (2) Systems based on fuzzy logic, (3) Systems based on pattern recognition, and (4) Systems based on neural networks.

#### 5.6.1 Systems Based on Yes/No Classification

The method commonly used by spectroscopists in practice is numerical analysis (Colthup, Daly, & Wiberley, 1990). Numerical analysis is primarily based on comparison between spectral data and reference values. Reference values are usually some regions like $frequency$ : $3615 \pm 5cm^{-1}$ or $strength$ : $1.000 \pm 0.300$. If spectral data are in certain regions, the answer of classification is yes; otherwise, the answer is no.

Most systems for interpreting infrared spectra use this method (Hasenoehrl, Perkins, & Griffiths, 1992; Puskar, Levine, & Lowry, 1986; Wythoff, Buck, & Tomellini, 1989). For example, in Wythoff's system, rules for comparing spectral data are in the following forms.

```
ANY PEAK(S)          FREQUENCY:1700-1707       STRENGTH:0.7-1.0
                     WIDTH:SHARP TO BROAD
      ANSWER -YES-
ACTION - ***
```

The advantage of these systems is that they are very easy to develop because they can directly use spectroscopic knowledge, and do not need further computation. However, the problem is that each of these systems is only applicable to a class of compounds, or pure compounds because in the case of seriously inaccurate spectral data, the reference values (regions) can not reflect the inaccuracy. For example, Hasenoehrl's system is only for distinguishing compounds containing at least one carbonyl functionality from other compounds, although the *RI* of the system is about 98% (naturally, the *RC* is not available), and Puskar's system is only for identifying hazardous substances.

In fact, spectroscopists also use qualitative analysis in some specific cases in addition to the formal spectroscopic knowledge, such as "*if the peaks in 600* $cm^{-1}$*- 900* $cm^{-1}$ *look like the peaks of benzene-rings, then the peaks in 3000* $cm^{-1}$*- 3100* $cm^{-1}$ *are quite likely to be created by a benzene-ring*". Unfortunately, the qualitative analysis was hardly applied to these systems since it can not be used in usual ways. In contrast, our system can successfully use the qualitative analysis like spectroscopists. The way of using it is the method proposed in this paper. As a result, our system is applicable to all compounds which exhibit high performance with respect to correctness.

#### 5.6.2 Systems Based on Fuzzy Logic

Since spectral data are always inaccurate, and the representation of spectroscopic knowledge is quite like that in fuzzy logic, some systems naturally use fuzzy logic or some techniques similar to fuzzy logic (Clerc, Pretsch, & Zurcher, 1986). In these systems, fuzzy intervals which are similar to the regions described in Section 5.6.1 are given for reference values, and memberships of inaccurate data are calculated on the basis of the degrees that the inaccurate data are in the fuzzy intervals. These systems are better than those described in Section 5.6.1 in some cases, but the degrees that inaccurate data are in fuzzy intervals do not necessarily reflect the possibility of the inaccurate data being the reference values. For example, in Figure 7, it is difficult to determine which peak is closer to the reference value only by considering the degrees that *peak a* and *peak b* are in the fuzzy interval.

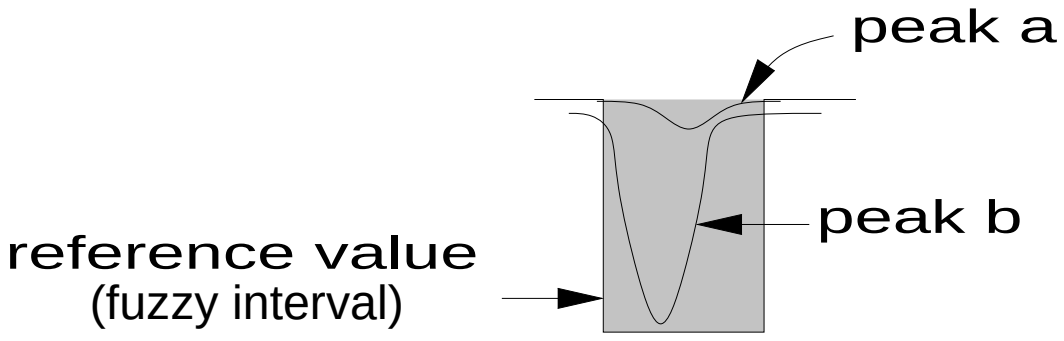

Figure 7: Two peaks in a fuzzy interval

However, by applying the method proposed in this paper, the above problem can be easily solved. As we discussed in Section 5.6.1, in practice spectroscopists also frequently use knowledge about correlations among peaks in addition to the formalizable spectroscopic knowledge. This kind of knowledge is essential to our method which enables us to use qualitative correlations among related data as evidence for the identification of inaccurate data.

We have compared the fuzzy method used by these systems with our method in Section 5.5. So far as we know, the *RC* of our system is the highest among the similar systems, and the *RI* of our system is higher than that of most of the systems.

#### 5.6.3 Systems Based on Pattern Recognition

Some systems use pattern recognition techniques to interpret infrared spectra (Jalsovszky & Holly, 1988; Sadtler, 1988), of which Sadtler is the most popular commercial system. The system compares known patterns with unknown ones, and determines the possibility of an unknown pattern being a known one by calculating the quantitative similarity or closeness between the two patterns.

Unlike fuzzy techniques, pattern recognition considers a group of data (i.e., a pattern) at the same time. However, pattern recognition is primarily based on quantitative analysis. We have discussed that in many cases especially when the inaccuracy of spectral data is not slight, qualitative features of spectral data are much more important than quantitative ones. For example, Figure 8 shows two simple cases. The difference between the two patterns in (a) is smaller than that in (b). From the viewpoint of Sadtler, the two patterns in (a) are closer than those in (b). However, the two patterns in (b) may be the same in some cases, while the two patterns in (a) may not be the same in any case. The reason is that the qualitative features (frequency positions of peaks) of the two patterns in (a) are different.

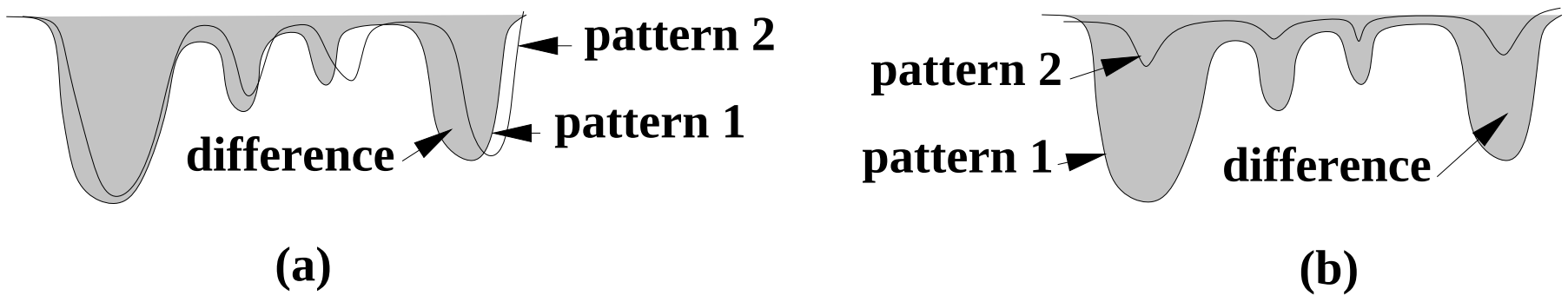

Figure 8: Quantitative differences between patterns

Because quantitative similarity and closeness are not always sound, most systems based on pattern recognition including Sadtler can not give concrete solutions. In general, the solutions of these systems are only a series of candidates from which users have to finally decide the possible one by themselves. It is difficult to compare these systems with ours because the solutions of these systems are quite loose, and neither the RC nor the RI is available. Sadtler, for example, usually gives the list of all known patterns associated with the values of quantitative differences between the unknown patterns and these known ones.

### 5.6.4 Systems Based on Neural Networks

Recently, neural networks have been applied to infrared spectrum interpreting systems (Anand, Mehrotra, Mohan, & Ranka, 1991; Robb & Munk, 1990). In Anand's system, a neural network approach is used to analyze the presence of amino acids in protein molecules. To this specific classification, the *RI* of Anand's system is about 87%, and the *RC* is not available. In Robb's system, a linear neural network model is developed for interpreting infrared spectra. The system is for general purpose like our system. Without prior input of spectrum-structure correlations, the *RC* of Robb's system is equal to 53.3%.

Although the *RC* and *RI* of our system are both higher than those of the two systems, we still think that using neural networks is very promising, especially when model training or system learning is a must. The research concerning applying neural networks to our system is left for the future.

## 6. Related Work and Discussion

Identifying inaccurate data has long been regarded as a significant and difficult problem in AI. Many methods and techniques have been proposed.

Fuzzy logic provides the mathematical fundamentals of representation and calculation of inaccurate data (Bowen, Lai, & Bahler, 1992; Negoita & Ralescu, 1987; Zadeh, 1978). Our method is primarily based on fuzzy theory. But compared with conventional fuzzy techniques, the advantages of our method include: (1) fuzzy intervals of inaccurate data are dynamically determined so that dynamic information can be used; (2) fuzzy intervals are based on qualitative features of data and qualitative correlations among related data so that the solutions are more robust. The limitation of our method is that when qualitative correlations among related data are not known in advance, the method degenerates to a conventional fuzzy method. For instance, if $SCF$ is unavailable, the two methods described in Section 5.5 become the same.

Pattern recognition provides the techniques for interpreting measured data in group (Jalsovszky & Holly, 1988). By pattern recognition methods, related data and connections among data can be considered. However, there are two preconditions which must be satisfied for complex data analysis by pattern recognition to be successful. The first precondition is that we have to obtain adequate data bases from which we can derive the patterns we need to recognize, and the second precondition is that we have to demonstrate that there are suitable metrics of similarity between patterns. When patterns explicitly exist, and measured patterns are not seriously noisy (e.g., fingerprint recognition), pattern recognition methods are effective. However, if patterns are not explicit, or patterns change irregularly which implies that there is not a stable metrics for determining the similarity between patterns (e.g., spectrum interpretation), our method is more practical and robust.

In identifying inaccurate data, the roles of "$d_i@R_j$" and "$R_j@MD$" are quite similar to the role of subjective statements or prior probabilities in other systems (Duda, Hart, & Nilsson, 1976; Shortliffe & Buchanan, 1975). However, the essential difference is that our method dynamically calculates the values of "$d_i@R_j$" and "$R_j@MD$" from qualitative correlations among related data so that it does not need many assumptions beforehand, and can avoid inconsistency in knowledge and data bases. Our method can also handle possibility propagation among inference networks. Readers may have noticed it from the process of considering the second kind of related data in spectrum interpretation (see Section 5.4.2).

When statistical samples are sufficient, or subjective statements can be consistently obtained, probabilistic reasoning methods can be applied to inaccurate data identification. When statistical samples of inaccurate data are not enough and consistent subjective statements are not available, our method is very effective.

Our ongoing research related to probabilistic reasoning is to consider the interaction among identified partial components. As we discussed before, spectroscopists frequently use the knowledge such as *"if $C_6H_6$ coexists with $CH_3$, then the peaks of $CH_3$ around 2900 $cm^{-1}$ may shift"*, or *"if -C-O-C- has been identified, then the strength of the peaks of $CH_3$ may change"*. Therefore, it is possible to update the possibilities of identified partial components by considering the interaction among them. Using probabilistic reasoning to analyze the effects among identified partial components would not only help us identify inaccurate data, but also provide us with the reason why the data are inaccurate. The research and experiments will be the subject of our sequel paper.

## 7. Conclusions

In this paper, we have presented a new method for identifying inaccurate data on the basis of qualitative correlations among related data. We first introduced a new concept called support coefficient function ($SCF$). Then, we proposed an approach to determining dynamic shift intervals of inaccurate data based on $SCF$, and an approach to calculating possibility of identifying inaccurate data, respectively. We also presented an algorithm for using qualitative correlations among related data as confirmatory or disconfirmatory evidence for the identification of inaccurate data. We have developed a practical system for interpreting infrared spectra by applying the proposed method, and have fully tested the system against several hundred real spectra. The experimental results show that the proposed method is significantly better than the conventional methods used in many similar systems. In this paper we have also described the system and the experimental results.

Briefly, our novel work includes:

1. A method which assumes an inaccurate data item to be a certain reference value on the basis of qualitative correlations between the inaccurate data item and all of its related data.
2. An algorithm which crystallizes the method.
3. A practical system which uses the algorithm to interpret infrared spectra.

## Acknowledgments

Thanks to the editors and anonymous reviewers of JAIR for their helpful comments and suggestions, and to Chunling Sui and Mitchell Bradt for proofreading the manuscript. This research was partially supported by Horiba Ltd., Kyoto, Japan, and the first author wishes to thank ASTEM Research Institute, Kyoto, Japan, where he worked as a researcher in 1991 - 1994.

## References

Anand, R., Mehrotra, K., Mohan, C. K., & Ranka, S. (1991). Analyzing Images Containing Multiple Sparse Patterns with Neural Networks. In *Proceedings of IJCAI-91, pp. 838-843.*

Bowen, J., Lai, R., & Bahler, D. (1992). Lexical Imprecision in Fuzzy Constraint Networks. In *Proceedings of AAAI-92, pp. 616-621.*

Clerc, J. T., Pretsch, E., & Zurcher, M. (1986). Performance Analysis of Infrared Library Search Systems. *Mikrochim. Acta [Wien], II, pp. 217-242.*

Colthup, L., Daly, H., & Wiberley, S. E. (1990). *Introduction to Infrared and Raman Spectroscopy.* Academic Press.

Dempster, A. P. (1968). A Generalization of Bayesian Inference. *Journal of the Royal Statistical Society, B-30, pp. 205-247.*

Duda, R. O., Hart, P. E., & Nilsson, N. J. (1976). Subjective Bayesian Methods for Rule-Based Inference Systems. In *Proceedings of National Computer Conference, pp. 1075-1082.*

Hasenoehrl, E. J., Perkins, J. H., & Griffiths, P. R. (1992). Expert System Based on Principal Components Analysis for the Identification of Molecular Structures from Vapor-Phase Infrared Spectra. *Journal of Anal. Chem., 64, pp. 656-663.*

Jalsovszky, G. & Holly, G. (1988). Pattern Recognition Applied to Vapour-Phase Infrared Spectra: Characteristics of vOH Bands. *Journal of Molecular Structure, 175, pp. 263-270.*

Negoita, C. V. & Ralescu, D. (1987). *Simulation, Knowledge-Based Computing, and Fuzzy Statistics.* Van Nostrand Reinhold Company.

Pearl, J. (1988). *Probabilistic Reasoning in Intelligent Systems: Networks of Plausible Inference.* Morgan Kaufmann Publishers.

Puskar, M. A., Levine, S. P., & Lowry, S. R. (1986). Computerized Infrared Spectral Identification of Compounds Frequently Found at Hazardous Waste Sites. *Journal of Anal. Chem., 58, pp. 1156-1162.*

Reiter, R. (1987). A Theory of Diagnosis From First Principles. *Artificial Intelligence, (87) 32, pp. 57-95.*

Robb, E. W. & Munk, M. E. (1990). A Neural Network Approach to Infrared Spectrum Interpretation. *Mikrochim. Acta [Wien], I, pp. 131-155.*

Sadtler Research Laboratories. (1988). *Sadtler PC Spectral Search Libraries, Product Introduction & User's Manual.*

Shafer, G. (1976). *A Mathematical Theory of Evidence.* Princeton Uni. Press.

Shortliffe, E. H. & Buchanan, B. G. (1975). A Model of Inexact Reasoning in Medicine. *Mathematical Biosciences, 23, pp. 351-379.*

Wythoff, B. J., Buck, C. F., & Tomellini, S. A. (1989). Descriptive Interactive Computer-Assisted Interpretation of Infrared Spectra. *Analytica Chimica Acta, 217, pp. 203-216.*

Zadeh, L. A. (1978). Fuzzy Set as a Basis for a Theory of Possibility. *Fuzzy Sets Syst., 1, pp. 3-28.*

Zhao, Q. (1994). An Efficient Method of Solving Constraint Satisfaction Problems in IR Spectrum Interpretation. In *Proceedings of the 2nd International Conference on Expert Systems for Development, pp. 165-170.*

Zhao, Q. & Nishida, T. (1994). A Knowledge Model for Infrared Spectrum Processing. In *Proceedings of the International Symposium on Information Theory and Its Applications, pp. 781-786.*